\documentclass[conference]{IEEEtran}

\usepackage[utf8]{inputenc} 
\usepackage[english]{babel}
\usepackage[T1]{fontenc}  
\usepackage{todonotes}
\usepackage{epsfig}           
\usepackage{subfigure}
\usepackage{graphicx}
\usepackage{setspace}
\usepackage{amsmath}
\usepackage{url}
\usepackage{amsfonts}
\usepackage{amssymb}
\usepackage{calligra}
\usepackage{eucal}
\usepackage[linesnumbered]{algorithm2e}
\usepackage{balance}
\usepackage[bottom]{footmisc}

\usepackage{lscape}
\usepackage{color}
\IEEEoverridecommandlockouts

\begin {document}

\title{An End-to-End Approach for Recognition of Modern and Historical Handwritten Numeral Strings}

\author{
    \IEEEauthorblockN{Andre G. Hochuli, Alceu S. Britto Jr.,\\ Jean P. Barddal}
	\IEEEauthorblockA{Graduate Program in Informatics (PPGIa)\\
	    Pontif\'icia Universidade Cat\'olica do Paran\'a\\
		Curitiba, PR - Brazil\\
		\{aghochuli, alceu, jean.barddal\}@ppgia.pucpr.br}
	\and
    \IEEEauthorblockN{Luiz  E. S. Oliveira}
	\IEEEauthorblockA{Department of Informatics (DInf)\\
	    Universidade Federal do Paran\'a\\
		Curitiba, PR - Brazil\\
		lesoliveira@inf.ufpr.br}
	\and
	\IEEEauthorblockN{Robert Sabourin}
	\IEEEauthorblockA{System Engineering Dept. (LIVIA)\\
	\'Ecole de Technologie Sup\'erieure\\
		Montreal, QC - Canada\\
		robert.sabourin@etsmtl.ca}

	}

\maketitle

\begin{abstract}
An end-to-end solution for handwritten numeral string recognition is proposed, in which the numeral string is considered as composed of objects automatically detected and recognized by a YoLo-based model. 
The main contribution of this paper is to avoid heuristic-based methods for string preprocessing and segmentation, the need for task-oriented classifiers, and also the use of specific constraints related to the string length. 
A robust experimental protocol based on several numeral string datasets, including one composed of historical documents, has shown that the proposed method is a feasible end-to-end solution for numeral string recognition. 
Besides, it reduces the complexity of the string recognition task considerably since it drops out classical steps, in special preprocessing, segmentation, and a set of classifiers devoted to strings with a specific length.  
\end{abstract}

\IEEEpeerreviewmaketitle


\section{Introduction}

Handwritten digit string recognition (HDSR) has been a subject of research over the past few decades. Additionally, in recent years, the information retrieval on historical documents has recovered attention to this field since plenty of digitalized historical document datasets were released \cite{ARDIS2019,OCR4ALL2019}.

To deal with the HSDR task, most of the proposed works segment the string into isolated digits and then apply a classifier capable of recognizing 10 classes $(0,\ldots,9)$. 
However, this approach becomes unfeasible in the presence of noise, broken digits, and touching digits. 

The literature provides several approaches to deal with the presence of touching digits. 
Most of the algorithms are based on contour and profile information to over-segment the numerical string, thus generating components that may represent a digit or parts of it. 
Next, a fusion method determines the best combination according to the \textit{a posteriori} probabilities.
Even though over-segmentation yields interesting recognition rates, its computational cost is elevated.

The alternatives resort to segmentation-free based methods
\cite{Choi99,Ciresan2008,Hochuli2018,Hochuli2018b,Aly2019} in which the string is recognized without the need for its prior segmentation into isolated digits. 
Such an approach has recovered the attention of the research community in the last years with the recent advances in deep learning. 
While over-segmentation methods demand (i) some specific strategy to generate segmentation cuts, (ii) a robust isolated digit recognizer, and (iii) a strategy for searching the best path among the generated segmentation hypothesis; segmentation-free approaches demand a significant amount of training data. It is worth to remark that, despite the fact these approaches boost the recognition rates, the background suppression and the detection of connected components remain a bottleneck since they are based on a set of heuristics.
         
Some of the aforementioned issues such as background suppression, noise, and touching components, are also related to the field of object recognition, which is maturing very rapidly. 
Thus, there is a plethora of deep learning end-to-end models available in the state-of-art \cite{FasterRCNN,YOLO2016,YOLO2017}. 
In this paper, we investigate the advances made in object recognition to push the frontiers of handwriting recognition.

When discussing object recognition, one aspect that is very often highlighted in the literature is the importance of the context \cite{divvala2009empirical}.Although the contextual information is more limited in HDSR, it also plays an important role, as demonstrated in \cite{Oliveira02b}. 

In this paper, we defend that handwritten digits can be seen as objects; hence a string of digits is a sequence of objects. 
The architecture used in this work is the YoLo \cite{YOLO2016}\cite{YOLO2017}, which provides appropriate features for the digit context. 
Furthermore, we demonstrate that when applicable, a synthetic dataset of strings mimicking real datasets can provide reliable contextual information, thus minimizing data annotation efforts. 

The main contribution of this work is an elegant end-to-end approach free of i) heuristic-based preprocessing, ii) heuristic-based segmentation, iii) multiple task-specific classifiers, and iv) constraints about the number of touching digits in the strings. 
In order to validate the proposed approach we present experiments on several datasets. 
First, we show the importance of the synthetic dataset created to provide contextual information. 
Second, we tested the limits of the proposed approach on very large strings (up to 20 digits). 
Then, we define three real-world applications to assess the method: (a) 11,585 numerical strings, ranging from 2 to 6 digits, extracted from forms of NIST SD19 \cite{NISTSD192016}, (b) the courtesy amount of bank checks of ORAND-CAR datasets \cite{ILSVRC15}, and finally, the recent released ARDIS dataset\cite{ARDIS2019}, composed of numerical strings extracted from Swedish historical documents. 

The results in the aforementioned datasets show that the proposed strategy compares favorably to the segmentation-based and segmentation-free approaches published in the state-of-art with the clear advantage of having a shorter pipeline that minimizes the presence of heuristics-based modules, such as preprocessing.

\section{Related Work}\label{related_work:sec}

To avoid the burden of over-segmentation, some authors have devoted efforts towards segmentation-free approaches. 
In this vein, one of the first strategies was proposed in \cite{Choi99}. 
To avoid the segmentation of touching pairs, the authors designed a framework based on 100 neural networks. 
Their approach achieves 95.3\% recognition rate of touching pairs extracted from NIST-SD19 \cite{NISTSD192016}. 
A decade later, the work of \cite{Ciresan2008} took advantage of Convolutional Neural Networks (CNNs).
Two CNNs were trained, one for isolated digits and one for touching pairs. 
The authors combined these two networks to recognize 3-digit strings of NIST database, achieving a recognition rate of 93.4\%. 
At that time, strings with three connected digits were not considered.
    
Taking advantage of the advances of deep learning \cite{LecunNature2015}, authors in \cite{Hochuli2018} introduced a segmentation-free approach to recognize digit strings of any size. 
In their work, the authors combined four CNNs into a Dynamic Selection (DS) scheme \cite{BRITTO2014}. The first CNN works as a high-level classifier that determines the size of the components, while the other three operate in a low-level by classifying 1-digit, 2-digit, and 3-digit components, respectively. 
This approach achieved the state-of-the-art for NIST-SD19 \cite{NISTSD192016} and TDP \cite{Ribas2013} datasets, surpassing segmentation-based and segmentation-free methods.

In spite of the good performance, this approach has bottlenecks. 
First, it is based on a hierarchical framework composed of heuristic-based preprocessing and four classifiers, which means different sources of errors. 
Second, the strategy recognizes strings of any size but limited to 3-digit touching. 
To mitigate some of these problems, in \cite{Hochuli2018b}, the authors reduced the number of classifiers by introducing a single classifier ($\mathcal{C}_{1110}$) to classify those 1110 classes ($0 \ldots 9, 00 \ldots 99, \mbox{ and } 000 \ldots 999$). Although these approaches achieve high recognition rates, they still embed complex pipelines, surrounded by heuristic processes, preprocessing modules, and fusion strategies. 

In this work, we argue that handwritten digit string recognition can benefit from the recent advances in the field of object recognition, where the main goal consists in detecting and recognizing a set of predefined classes of objects in a given input image. 
Until the last decade, a classical approach was the sliding window based methods and its variants \cite{Felzenszwalb2008,Lampert2008, Felzenszwalb2010}.
This kind of approach repurposes a classifier trained with handcrafted features at several spatial locations of the image. 
A drawback is the elevated number of windows required to search over multiple scales and aspect ratios, which lead to increased computational costs. 

Major breakthrough happened when large-scale datasets \cite{ILSVRC15, MSCOCO} arose and GPUs became popular, enabling the efficient training of deep neural networks in the ILSVRC 2012 \cite{ILSVRC15}. 
Consequently, this field had recovered the attention of the research community, and several deep learning methods were proposed and set the new standards for the area.
     
One of the first successful approaches was the Region-based Convolutional Network (R-CNN) proposed by Girshick et al. \cite{RCNN}. 
This architecture first extracts region proposals from image space using the selective search algorithm \cite{SelectiveSearch}. 
Then, each region is warped to a fixed size, and a CNN extracts features. 
Finally, an SVM classifier determines a class, and a bounding-box regressor refines the locations. 
The necessity to extract features of each warped region is the main drawback of this strategy, which is computationally expensive. 
    
To overcome this obstacle, SPPnet \cite{SPPnet} and Faster-RCNN \cite{FasterRCNN} have been proposed. 
These models predict region proposals direct over feature maps. 
A spatial pooling layer was introduced to produce fixed-length representations (wrapping at feature level).
Although those strategies speed up the whole process, they still rely on a handcrafted region proposal method.
The work of \cite{MaskRCNN} then presented a region proposal network (RPN), which implicit produces the candidate locations. 
The features available in the last convolutional layer are used for both (i) region proposal and (ii) region classification. 
    
The above approaches still handle a two-stage pipeline once they require a region proposal strategy, regardless of whether it is implicit or not. 
A cleverer alternative was proposed in \cite{YOLO2016}\cite{YOLO2017} with the YoLo architecture, in which the authors introduced a regression-based approach that encapsulates all stages into a single network. 
With a single forward pass, the network provides bounding box locations and class probabilities. 
An important aspect of YoLo is that it can encode context and appearance from the neighborhood of objects, which is a very important feature for implicit digit segmentation.
In light of this, the YoLo model was selected as the object detector for this work. 
 
\section{Problem Statement}\label{yolo4digits:sec}	 

As stated before, the traditional approaches group foreground pixels into connected components and classifies them individually. 
The issue is that this approach provides a local view of the problem, thus ignoring contextual information. 
Without this valuable information, the algorithms suffer from the presence of noise and touching digits.    

An end-to-end approach addresses this problem in a global manner. 
Deep learning models can learn the interaction between digits in the context of an image, which contains noise, touching, overlapping, and broken digits. 
Therefore, the end-to-end approach features a quite short pipeline: the object detector $\mathcal{D}$ (Section \ref{yolo2:sec}) receives as input an image $I$ containing $n$ digits (objects) and produces as output the location (bounding boxes) and the digit classes $[0, \ldots, 9]$ associated with an estimation of the \textit{a posteriori} probability. 
Considering that the input image $I$ may contain $n$ connected components, the most probable interpretation of the written amount $M$ is given by Equation \ref{eq:prob2}, where $\omega_i = \{0 \ldots 9\}$ and $x_i$ stands for the digit candidates.

\begin{equation}
    P(M|I) = \prod_{i=1}^n P(\omega_j|x_i)
\label{eq:prob2}
\end{equation}

The simplicity of the proposed method is clear when we compare both systems. 
First, the end-to-end solution cuts off the preprocessing module to detect connected components. 
Next, it is unnecessary to train specialized classifiers or to design a fusion method to combine them. 
Since the digit string is a string of objects, the end-to-end approach imposes no constraints over the number of touching components. 
On the other hand, this kind of approach demands a considerable amount of data to fine-tune the network and hyper-parameters. 
The Section \ref{nistexp:sec} discusses a dataset tailored for efficiently learning this type of representation via synthetic strings.

\section{YoLo for Digit String Recognition}\label{yolo2:sec}

\begin{figure}[!h]
    \centering
    \epsfig {file=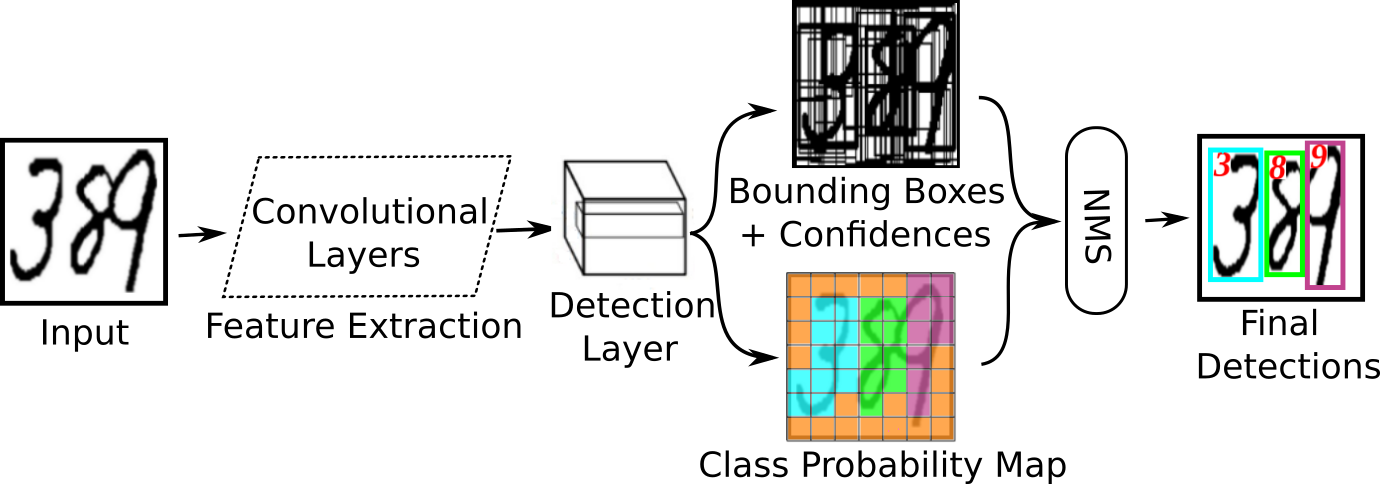, width=.48\textwidth}
    \caption{YoLo Framework: The system divides the image into a grid and for each cell predicts bounding boxes and classes.}
    \label{yoloframework:fig}
\end{figure}

The YoLo \cite{YOLO2016} is a general-purpose object detector framework trainable in an end-to-end manner. Using a single neural network and looking at the entire image, it can predict bounding boxes and classes with a single forward pass instead of applying the model at every location, such as traditional sliding window or region purpose-based methods \cite{FasterRCNN}. 
The framework is illustrated in Figure \ref{yoloframework:fig}.

First, the convolutional layers (Section \ref{yolo_layers:sec}) extract features from the entire image, and the detection layer divides the image into a grid. 
Next, each grid cell predicts bounding box coordinates with a confidence level that this box encloses an object.
To help the network to learn how to predict good bounding boxes, handpicked anchor boxes (Section \ref{anchors:sec}) are previously defined. 
Moreover, it provides class probabilities for those cells that belong to an object. 
Finally, to mitigate confusion among overlapped boxes, the well-known Non-Maximum Suppression (NMS) algorithm is used.

\subsubsection{Convolutional Layers}\label{yolo_layers:sec}

The architecture of the network (a.k.a. Darknet) is composed of 19 convolutional layers and 5 max-pooling layers. 
The input resolution of the Darknet is $416 \times 416$ \cite{YOLO2016}. 
However, given that strings of digits are usually wider than higher, we have used an initial input size of $128 \times 256$ $(height \times width)$ for initial training.
The input size changes during training, as every 10 batches, the network randomly chooses a new image dimension size, and the training is resumed. 
This forces the network to learn to predict well across a variety of input dimensions.

Section \ref{input-image-size:sec} depicts experiments that during recognition, the input size can be defined as a function of the testing input image. 
Since YoLo looks at the whole input, it implicitly encodes contextual information about objects and their neighborhood.

\subsubsection{Anchors}\label{anchors:sec}
The dimensions of anchor boxes were defined by authors using samples of Imagenet dataset\cite{ILSVRC15}, which is composed of 1000 classes of real-life objects. 
To optimize anchors for digits, we performed a $k$-means clustering over 10,000 ground-truth bounding boxes from training samples. 
The resulting anchors have 0.5, 0.6, and 1.0 aspect ratios.

\subsubsection{Training}\label{detectmodule:sec}

Training is performed using Stochastic Gradient Descent (SGD) and back-propagation with mini-batches of 64 instances, a momentum factor of 0.9, and a weight decay of $5 \times 10^{-4}$. 
The initial learning rate was set to $10^{-3}$, which allow weights to quickly fit the long ravines in the weight space, after which it is reduced over time (until $5 \times 10^{-4}$) to make the weights fit the sharp curvatures. 
The sum-squared error is used as a loss function to optimize detection and classification simultaneously.

In the present work, regularization was implemented through early-stopping, which prevents overfitting as training is interrupted as the performance of the network on a validation set deteriorates. 
The models were trained using NVidia GeForce Titan X and V GPUs\footnote{All trained classifiers are available for research purposes at https://web.inf.ufpr.br/vri/databases-software/touching-digits/}.

\section{Experiments}\label{sec:Experiments}

This section reports the assessment of our proposal in different scenarios.
First, we discuss the impacts of contextual information during the learning process, as YoLo models are trained with isolated digits and strings composed of single and touching digits. 
Next, we perform an experiment to find out which input image size maximizes the performance of the proposed model. 
Finally, our end-to-end approach is applied to different real-world applications: (a) strings extracted from forms of NIST SD19, (b) the courtesy amount of bank checks of ORAND-CAR dataset and finally, (c) numeral strings extracted from a Swedish historical dataset (ARDIS).

\subsection{The importance of contextual information}\label{context_learning:sec}

Throughout this paper, we have advocated the importance of the context for recognizing digits as objects. 
At this point, one may ask if it is really necessary as chaining the recognition of isolated digits has been used for the last three decades with reasonable success. 

\begin{figure}[!h]
    \begin{center}
        \subfigure[] {\epsfig {file=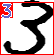, scale=0.55}}
        \hspace{0.7cm}    
        \subfigure[] {\epsfig {file=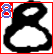, scale=0.55}}
        \hspace{0.7cm}    
        \subfigure[] {\epsfig {file=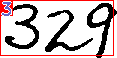, scale=0.55}}
        \hspace{0.7cm}
        \subfigure[] {\epsfig {file=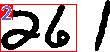, scale=0.55}}                                
        \caption{Outputs of YoLo trained with isolated digits only: (a) and (b) represent correct detections of isolated digits, while (c) and (d) are mis-detections on digit strings caused by the lack of context learning.}
        \label{context_learning:fig}
    \end{center}
    \vspace{-3mm}
\end{figure}

We trained the YoLo model using 197k samples of isolated digits from the \textit{hsf\_0123} folder of the NIST SD19 dataset. As a result, the model performs quite well recognizing isolated digits, achieving a performance around 99\% on a dataset composed on 23,621 isolated digits from the \textit{hsf\_7} folder of NIST SD19. However, when digit strings are presented, the model collapses in terms of both localization and recognition. 
The model tends to find only one component per image since they were trained with isolated objects. 
Some detection samples are illustrated on Figure \ref{context_learning:fig}.

\begin{figure}[!h]
    \begin{center}
        \subfigure[] {\epsfig {file=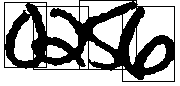, scale=0.3}}
        \hspace{0.7cm}    
        \subfigure[] {\epsfig {file=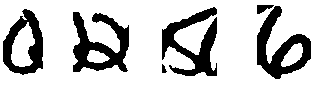, scale=0.3}}
            
        \caption{(a) Ground truth for a 4-digit string (0256) and (b) Shape of digits impacted by its neighbors.}
        \label{training:fig}
    \end{center}
    \vspace{-7mm}
\end{figure}

Besides, the shape of the digits may be severely affected by its neighbors. 
Consider for example the string depicted in Figure \ref{training:fig}a. 
Considering the ground-truth bounding boxes, the shape of some digits are quite different from those observed in the isolated digit datasets, especially those in the middle of the string. 
This is why learning from strings rather than isolated digits is important. 

\subsection{Input Image Size}\label{input-image-size:sec} 
As aforementioned, the images have been resized to $128 \times 256$ $(height \times width)$ for training. However, since YoLo changes the input size every few iterations during training, the network is able to recognize testing images of different sizes. 
The question is how to properly resize the testing input image to maximize the network's performance. 
This is a relevant issue since the image width may vary considerably according to the number of digits in the string. 
For instance, a 20-digit string is considerably longer than a 2-digit string, and thus, resizing both to $128 \times 256$ is not reasonable. 

To address this issue, we performed an experiment on 5,000 strings ranging from 2 to 20 digits, which were synthetically created by concatenating isolated digits from NIST SD19. 
For each string length, we tested the input image width in the following range: $[128,256,\ldots,1152,1280]$; whereas the image height is 128. 
Table \ref{differentinputsize:tab} summarizes the image input size that maximizes the recognition rate according to string length. 

\begin{table} [!h]
    \caption {Image input size that maximizes the recognition rate for each string length}
    \begin{center}
        \scalebox{0.85}{
        \begin{tabular}{cccc}
            \hline

            String     & Average      &Input Image  & Recognition       \\
            Length  & String Width &Size ($II_w$)  &Rate (\%)   \\
                     & ($S_w$)      &$(128 \times w)$                  &   \\ \hline
            2 & 75  &  128 & 98.6  \\
            6 & 228 &  384 & 97.6  \\
            10 &381 &  640 & 94.8  \\
            14 &524 &  896 & 91.0  \\
            20 &750 &  1280 & 89.6  \\ \hline
        
        \end{tabular}
        }
        \vspace{-5mm}
        \label{differentinputsize:tab}
    \end{center}
\end{table}

From Table \ref{differentinputsize:tab}, we can notice that there is a quasi-linear relation between the average string width of the testing images\footnote{The number of pixels may vary depending on the image resolution. In this work, all the images were acquired in 300dpi.} and the best input size for the YoLo. 
As a result, we derived Equation \ref{retina:eq} to compute the input size width of the YoLo based on the width of the testing image. 
This rule is used for all experiments reported in this paper.

\begin{equation}
II_w = \left \{ \begin{array}{ll}
                             128                 & \mbox{for } S_w \leqslant 75 \\
                              S_w \times 1.70    & \mbox{otherwise }    \\
        \end{array}
\right.
\label{retina:eq}
\end{equation}

\begin{figure*}[!h]
        \begin{center}
            \subfigure[] {\epsfig {file=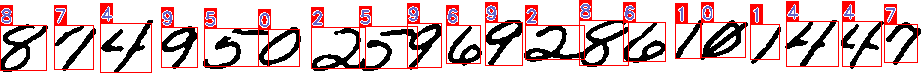, width=0.9\textwidth, height=1cm}}
            \caption{20-digit strings correctly recognized by the proposed approach.}
            \label{longstrings:fig}
        \end{center}
        \vspace{-5mm}
    \end{figure*}

Figure \ref{longstrings:fig} shows examples of 20-digit strings recognized by the system using the aforementioned rule. 
These corroborate the efficiency of the adopted resizing strategy and show that the proposal performs well even for long strings composed of broken, overlapping, and touching digits.

\subsection{NIST SD19 Strings}\label{nistexp:sec}
    
The experiments using real-world strings are based on 11,585 numeral strings extracted from the \textit{hsf\_7} series and distributed into five classes: 2\_digit (2,370), 3\_digit (2,385) 4\_digit (2,345), 5\_digit (2,316), and 6\_digit (2,169) strings, respectively. 
The strings were cropped from original samples leaving a border of 5 pixels.
These data exhibit different problems such as touching and fragmentation, and they were also used as a test set in \cite{Hochuli2018, Oliveira02b, Alceu2001, Liu2004, Oliveira2004, Sadri2007, Gattal2017}. 
It is important to mention that at any moment, the writers of \textit{hsf\_7} series were not used for training.

To provide a sufficient amount of data to learn the representation, we created a synthetic dataset composed of 365,000 numerical strings ranging from 2- to 6-digits, containing isolated and touching components\footnote{All the synthetic data is available upon request for research purposes at \url{https://web.inf.ufpr.br/vri/databases-software/touching-digits/}}. 
The rationale of this strategy is to create a dataset with contextual information about the neighborhood of isolated and touching digits that is not present in traditional learning with single-digit samples. 

\begin{figure}[!h]
    \begin{center}
        \subfigure[] {\epsfig {file=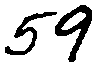, width=1.8cm, height=0.7cm}}
        \hspace{0.5cm}
        \subfigure[] {\epsfig {file=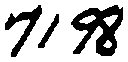, width=2cm, height=0.7cm}}
        \hspace{0.5cm}            
        \subfigure[] {\epsfig {file=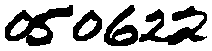, width=2.2cm, height=0.7cm}}
        \caption{Synthetic data representing numerical strings ranging from 2 to 6 digits.}
        \label{synthdataset:fig}
    \end{center}
    \vspace{-5mm}
\end{figure}

An in-depth discussion about this is presented on Section \ref{context_learning:sec}. 
The strings are built by concatenating isolated digits of the \textit{hsf\_{0123}} folder of NIST SD19 \cite{NISTSD192016} through the algorithm described by Ribas et al. in \cite{Ribas2013}. 
Figure \ref{synthdataset:fig} depicts some examples. 
To avoid potential biases, we used the information of the authors available on the NIST SD19, such that digits from different authors were used exclusively for training, validation, and testing.

Table \ref{summarystring:tab} summarizes the results for this experiment. 
We see that there are two sources of errors: detection (when the digit is mispositioned or not detected at all) and classification (when the digit is correctly detected, but it is misclassified).

\begin{table} [!h]
    \caption {Accuracy (\%) for the NIST Strings}
    \begin{center}
        \scalebox{0.85}{
        \begin{tabular}{ccccc}
            \hline
            Length & Samples & Accuracy (\%) & \multicolumn{2}{c}{Error (\%)} \\ \cline{4-5} 
            &  &  & Classification & Detection \\ \hline
            2 & 2370 & 98.57 & 1.39 & 0.04 \\
            3 & 2385 & 97.61 & 2.32 & 0.08 \\
            4 & 2345 & 97.10 & 2.56 & 0.34 \\
            5 & 2316 & 96.50 & 2.59 & 0.91 \\
            6 & 2169 & 95.80 & 3.14 & 1.06 \\ \hline
            & Average & 97.12 & 2.40 & 0.49 \\ \hline
        \end{tabular}
        }
        \label{summarystring:tab}
    \end{center}
    \vspace{-5mm}
\end{table}

\begin{figure}[!h]
        \begin{center}
            \subfigure[] {\epsfig {file=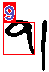, scale=0.5}}
            \hspace{0.2cm}    
            \subfigure[] {\epsfig {file=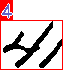, scale=0.7}}
            \hspace{0.2cm}                
            \subfigure[] {\epsfig {file=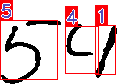, scale=0.5}}
            \hspace{0.2cm}    
            \subfigure[] {\epsfig {file=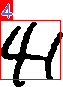, scale=0.5}}    
            \hspace{0.2cm}    
            \subfigure[] {\epsfig {file=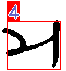, scale=0.5}}    
            
            \subfigure[] {\epsfig {file=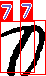, scale=0.5}}
            \hspace{0.2cm}    
            \subfigure[] {\epsfig {file=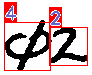, scale=0.5}}
            \hspace{0.2cm}    
            \subfigure[] {\epsfig {file=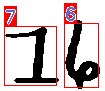,  height=1.2cm}}
            \hspace{0.2cm}    
            \subfigure[] {\epsfig {file=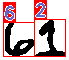, scale=0.5}}
            \hspace{0.2cm}    
            \subfigure[] {\epsfig {file=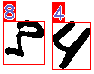, scale=0.5}}    
            
            \caption{Missed predictions: (a) to (f) representing missed detections and (g) to (j) representing misclassifications}
            \label{misd:fig}
        \end{center}
        \vspace{-2mm}
    \end{figure}

Regarding the detection errors, which the average error rate is below 1\%, we observed that most of the problems are related to the digit ``1''. 
The problem occurs when i) the height has a variance with his neighbor (Figure \ref{misd:fig}a), or ii) the slant of the image is big (Figure \ref{misd:fig}b). 
In these cases, the digits ``1'' were undetected. 
Another source of errors is the digit ``4'' (very often related to the digit ``1''). 
In these cases, sometimes the model detects two objects (``4'' and ``1'' ) in the digit ``4'' (Figure \ref{misd:fig}c) and sometimes just the digit ``4'' is detected, missing the digit ``1' (Figure \ref{misd:fig}d). 
Finally, we have observed few samples similar to under-segmentation \ref{misd:fig}e) and over-segmentation (Figure \ref{misd:fig}f).

With respect to the misclassification, Table \ref{summarystring:tab} shows an average error rate of 2.4\%. 
Performing an error analysis, we noticed that most confusions are related to the variability of the handwriting. 
From Figure \ref{misd:fig}g to Figure \ref{misd:fig}i are illustrated some common mistakes involving classes ``0'' and ``1''. 
In such cases, these handwriting styles are poorly represented in the training set (Figure \ref{misd:fig}j).

Table \ref{comparisonstrings:tab} compares the accuracy (\%) of several systems reported in the literature on NIST-SD19. 
For the sake of completeness, we replicate the results compiled by the authors in \cite{Hochuli2018}. 
The works presented in \cite{Alceu2001}, \cite{Oliveira02b}, and \cite{Oliveira2004} use different segmentation (implicit and explicit) and classification strategies, such as Hidden Markov Models, Multi-layer Perceptrons and Support Vector Machines.
    
\begin{table} [!t]
    \caption {Comparison of the accuracy (\%) on NIST SD19}
    \begin{center}
        \scalebox{.77}{
            \begin{tabular}{|c|cccccccccc|} 
                \multicolumn{1}{c}{\rotatebox{90}{Length}} &
                \multicolumn{1}{c}{\rotatebox{90}{Samples}} &
                \multicolumn{1}{c}{\rotatebox{90}{\cite{Alceu2001}}} &
                \multicolumn{1}{c}{\rotatebox{90}{\cite{Oliveira02b}}} &
                \multicolumn{1}{c}{\rotatebox{90}{\cite{Oliveira2004}}} &
                \multicolumn{1}{c}{\rotatebox{90}{\cite{Sadri2007}}} &
                \multicolumn{1}{c}{\rotatebox{90}{*\cite{Sadri2007}}} &
                \multicolumn{1}{c}{\rotatebox{90}{\cite{Gattal2017}}} &
                \multicolumn{1}{c}{\rotatebox{90}{\cite{Hochuli2018}}} &
                \multicolumn{1}{c}{\rotatebox{90}{\cite{Aly2019}}} &
                \multicolumn{1}{c}{\rotatebox{90}{\bf{YoLo}}} \\ \hline

                2 & 2370 & 94.8 & 96.8 & 97.6 & 95.5 & 98.9 & 99.0 & 97.6 & 98.8 & \textbf{98.6}\\ 
                3 & 2385 & 91.6 & 95.3 & 96.2 & 91.4 & 97.2 & 97.3 & 96.2 & 96.4 & \textbf{97.6}\\ 
                4 & 2345 & 91.3 & 93.3 & 94.2 & 91.0 & 96.1 & 96.5 & 94.6 & 95.0 & \textbf{97.1}\\
                5 & 2316 & 88.3 & 92.4 & 94.0 & 88.0 & 95.8 & 95.9 & 94.1 & 95.4 & \textbf{96.5}\\
                6 & 2169 & 89.0 & 93.1 & 93.8 & 88.6 & 96.1 & 96.6 & 93.3 & 95.0 & \textbf{95.8}\\ \hline
                \multicolumn{2}{|c}{\textbf{Average}} & 91.0 & 94.2 & 95.2 & 90.9 & 96.8 & 97.1 & 95.2 & 96.1 & \textbf{97.1} \\\hline 
                
            \end{tabular}
        }
        \vspace{-5mm}
        \label{comparisonstrings:tab}
    \end{center}
\end{table}

The work presented by Sadri et al. \cite{Sadri2007} is reported in two columns. 
The authors proposed a system based on over-segmentation, in which they used a genetic algorithm to optimize their segmentation algorithm. 
As pointed out in \cite{Hochuli2018}, the second set of experiments (marked with * in Table \ref{comparisonstrings:tab}) is somehow biased since the heuristics were defined using a subset of the testing set. 
Good performance was also reported by Gattal et al. \cite{Gattal2017}, yet, their segmentation thresholds have been adjusted on the test set.

Finally, a straightforward comparison is possible with the segmentation-free method proposed in \cite{Hochuli2018} and recently improved by \cite{Aly2019}, in which the fusion rule was eliminated by a cascade architecture of PCA-SVMNet classifiers, however, keeping the preprocessing steps and specific-task classifiers. 
As discussed in Section \ref{yolo4digits:sec}, the proposed approach improves accuracy whilst cutting off all the heuristics used for preprocessing, the necessity of training several deep learning models, and the parameter used in the fusion strategy. 

\subsection{ORAND-CAR Datasets}\label{orand-car:sec}

\begin{figure}[!b]
    \begin{center}
        \subfigure[76210] {\epsfig {file=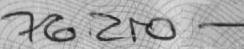,  width=0.15\textwidth, height=0.9cm}}
        \hspace{0.5cm}            
        \subfigure[60000] {\epsfig {file=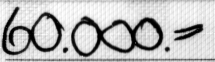, width=0.15\textwidth, height=0.9cm}}
        \hspace{0.5cm}

        \caption{Sample data of (a) Car-A and (b) Car-B datasets.}
        \label{orand-cvl:fig}
    \end{center}
    \vspace{-5mm}
\end{figure}

This experiment was performed on real-world datasets built for the ICFHR 2014 challenge on HDSR \cite{Diem2014}.
The ORAND-CAR-2014 consists of digit strings of the courtesy amount recognition (CAR) field extracted from real bank checks with a resolution of 200 dpi.
Besides the traditional challenges in handwriting such as noise, broken, and touching digits, this dataset presents samples with background and currency symbols such as `\#',  `\$', dots, commas, and dashes. 
It includes variation in size as well as writing style. 
This database poses new challenges to the community since it is harder than other datasets, especially in terms of variance in writing style. 
Table \ref{orand-cvl-info:tab} shows the amount of data used for training and testing and some examples are depicted in Figure \ref{orand-cvl:fig}.

\begin{table}[!b]
    \begin{center}
        \caption{Distribution of Orand-Car}
        \scalebox{0.85}{
        \begin{tabular}{c|ccc|ccc}
            \hline
            &  & Car-A &  &  & Car-B \\
            Length & Train & Val & Test & Train & Val & Test \\ \hline
            2 & 17 & 5 & 36 & 0 & 0 & 0 \\
            3 & 176 & 28 & 387 & 0 & 0 & 0  \\
            4 & 633 & 71 & 1425 & 60 & 3 & 5  \\
            5 & 819 & 84 & 1475 & 1080 & 120 & 69  \\
            6 & 127 & 18 & 363 & 1432 & 167 & 1241  \\
            7 & 27 & 2 & 87 & 127 & 10 & 1452   \\
            8 & 1 & 1 & 11 & 1 & 0 & 157  \\
            9 & 0 & 0 & 0 & 0 & 0 & 2  \\\hline
            Total & 1800 & 209 & 3784 & 2700 & 300 & 2926  \\ \hline
        \end{tabular}
        }
        \label{orand-cvl-info:tab}
    \end{center}
    \vspace{-2mm}
\end{table}

Once this dataset has different handwriting styles than NIST SD19, the use of models constructed with synthetic data use so far would provide unreliable results. 
Thus, we have trained all models using the data described in Table \ref{orand-cvl-info:tab}, since it is the protocol suggested in the competition. 
We kept the training parameters unchanged, following the description provided in Section \ref{yolo2:sec}. 
To provide sufficient information to object-detection approach, we have annotated the digits bounding-boxes (ground-truths) of each training sample\footnote{The annotated dataset is available upon request for research purposes at https://web.inf.ufpr.br/vri/databases-software/touching-digits/}. 
This laborious task was necessary since most of the samples have a complex background, noise, and symbols, which is difficult to reproduce synthetically.

\begin{table}[!b]
    \caption {Comparison of the accuracy (\%) on Orand-Car datasets Reported by \cite{Xu2018}}
    \begin{center}
        \scalebox{0.85}{
        \begin{tabular}{cccc}
            \hline
            Methods & CAR-A & CAR-B \\ \hline
            Tebessa I\cite{Diem2014} & 37.05 & 26.62 \\
            Tebessa II\cite{Diem2014} & 39.72 & 27.72  \\
            Hochuli et al.\cite{Hochuli2018} & 50.10 & 40.20 \\
            Singapore\cite{Diem2014} & 52.30 & 59.30  \\
            Pernanbuco\cite{Diem2014} & 78.30 & 75.43 \\
            Beijing\cite{Diem2014} & 80.73 & 70.13 \\
            CRNN\cite{Shi2017-CRNN} & 88.01 & 89.79 \\
            Saabni\cite{Saabni2016}$*$ & \multicolumn{2}{c}{85.80} \\
            ResNet-RNN\cite{Zhan2017} & 89.75 & 91.14 \\
            ResNet-RNN\cite{Xu2018} & 91.89 & 93.79 \\        
            \textbf{YoLo} & \textbf{96.20} & \textbf{96.80} \\ \hline            
            \multicolumn{3}{r}{$*$ Unified CAR-A and CAR-B datasets} & \\
            
        \end{tabular}
    }
                
        \label{orand-cvl:tab}    
        
    \end{center}            
\end{table}

The performance of the end-to-end approach on the test set is presented in Table \ref{orand-cvl:tab}. 
Differently from the other benchmarks, where the dynamic selection approach \cite{Hochuli2018} performed quite well, in these experiments, it struggled mostly because of its heuristic-based preprocessing module. 
Since ORAND-CAR provides a hard background and currency symbols, the preprocessing module collapsed in detecting the connected components. 

\begin{figure}[!t]
    \begin{center}
        \subfigure[] {\epsfig {file=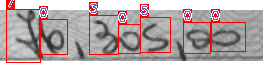, height=1.1cm}}
        \hspace{0.3cm}    
        \subfigure[] {\epsfig {file=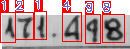, height=1.1cm}}
        

        \caption{Missed predictions of YoLo for ORAND dataset: (a) `7630500' as `7030500' and (b) `171448' as `121498'}
        \label{YoLo-orand:fig}
    \end{center}
    \vspace{-5mm}
\end{figure}

A remarkable performance was achieved by the YoLo, which reveals the robustness of the model in the task of encoding context, noise, and background.
The ORAND dataset also provides challenges in terms of overlapped digits, handwriting variability, and different aspect ratios that severely impact the performance of models.
Figure \ref{YoLo-orand:fig} depicts these issues.

\subsection{ARDIS Historical Dataset }\label{Ardis:sec}

The Swedish dataset of historical handwritten digits \cite{ARDIS2019} is composed of 4-digit image-based strings  (years) extracted from 15,000 Swedish church records, available in the Arkiv Digital Sweden (ARDIS).
The strings were written by different priests with various handwriting styles in the nineteenth and twentieth centuries, with different dip pens, and the alphabets are scripted in various sizes, directions, widths, arrangements, and measurements, which impose different challenges when compared to modern datasets.
The dataset also provides additional information about the city and the book type (category) that the image was collected.
    
\begin{figure}[!b]
        \begin{center}
            \subfigure[] {\epsfig {file=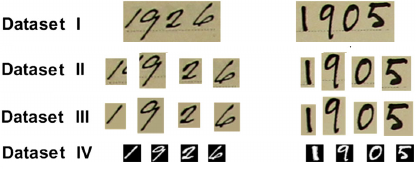,  width=0.45\textwidth, height=2cm}}
            \hspace{0.5cm}    
            \subfigure[] {\epsfig {file=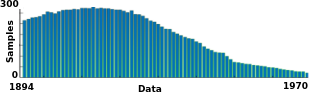, width=0.48\textwidth, height=2cm}}            
            
            \subfigure[] {\epsfig {file=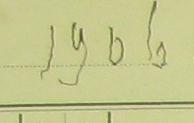, height=0.9cm}}
            \hspace{0.15cm}
            \subfigure[] {\epsfig {file=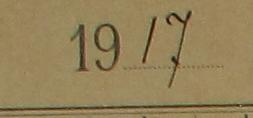, height=0.9cm}}
            \hspace{0.15cm}
            \subfigure[] {\epsfig {file=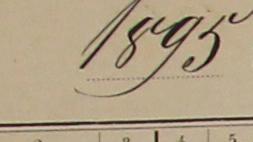, height=0.9cm}}
            \hspace{0.15cm}
            \subfigure[] {\epsfig {file=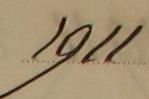, height=0.9cm}}                                    

            \caption{Ardis Dataset: (a) Data representation of each dataset, (b) the class distribution for strings (years) and (c)-(f) representing the variety in handwriting styles and backgrounds.}

            \label{ardis:fig}
        \end{center}
        \vspace{-5mm}
    \end{figure}

The dataset consists of one digit string dataset and three single-digit datasets (Figure \ref{ardis:fig}a). 
The digit strings dataset includes 10,000 samples in RGB color space, whereas, the other datasets contain 7,600 single-digit images in different color spaces. 
Figure \ref{ardis:fig}b depicts the string class distribution. 
    
\subsubsection{Single Digit Benchmark} 

The authors proposed a benchmark for single digits of Dataset IV reporting the accuracy of several machine learning models using different training protocols. 
The results are reported in Table \ref{ardis-single-results:tab}. 
In Case I, the models were trained with samples of the MNIST dataset and tested on ARDIS. 
For Case II, the samples of the USPS dataset are used for training. 
These two scenarios show how different is the representation of modern datasets from historical ones provided from ARDIS since the handwritten style is quite different (cursive gothic, high variance in the size and slant). 
In Case III, the ARDIS dataset is used for training and testing, while in Case IV, the MNIST and ARDIS training and test sets are merged to evaluate the generalization of the model. The YoLo surpasses the models of benchmark in all scenarios, even that it has digit localization as an additional task. In the worst performances, cases I and II, most of errors are related to the poorly representation of the handwritten style of the historical dataset (ARDIS) that is not present in modern ones (MNIST/USPS). The dynamic selection proposed by \cite{Hochuli2018} suffers in the generalization of length and 1-digit classifier.

    \begin{table}[!t]
        \caption{Benchmark for ARDIS Dataset IV (single digits) of models on different trainning protocols (reported by \cite{ARDIS2019})}
        \begin{center}
        \scalebox{0.85}{
            
        \begin{tabular}{c|cccc}
            \hline
            & \multicolumn{4}{c}{\textbf{Accuracy (\%)}} \\
            \textbf{Method} & \textbf{Case I} & \textbf{Case II} & \textbf{Case III} & \textbf{Case IV} \\             \hline
            \textbf{YoLo} & \textbf{87.60} & \textbf{64.10} &\textbf{ 99.70} & \textbf{99.27} \\        
            {Hochuli et al.\cite{Hochuli2018}} & 67.20 & 51.90 & 83.30 & 60.55 \\        
            {CNN} & 58.80 & 35.44 & 98.60 & 99.34 \\        
            {HOG-SVM} & 56.18 & 33.18 & 95.50 & 98.08 \\
            {RNN} & 45.74 & 28.96 & 91.12 & 96.74 \\    
            {kNN} & 50.15 & 22.72 & 89.60 & 96.63 \\
            {SVM} & 43.40 & 30.62 & 92.40 & 96.48 \\
            {Random Forest} & 20.12 & 17.15 & 87.00 & 93.12 \\
            \hline
        \end{tabular}
        }
        \end{center}
        \label{ardis-single-results:tab}
        \vspace{-5mm}
    \end{table}

\subsubsection{Digit String Benchmark} 
    
    Besides a high-quality analysis for single digits, the authors did not propose a benchmark for digit strings (years). 
    Thus, we evaluate the performance of the proposed end-to-end solution on digit strings (Dataset I).
    Moreover, we proposed a benchmark protocol for this dataset. 
    
\begin{figure}[!h]
        \begin{center}
            \subfigure[1897 to 1986] {\epsfig {file=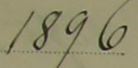, width=1.9cm, height=0.8cm}}
            \hspace{0.02cm}            
            \subfigure[1909 to 190910] {\epsfig {file=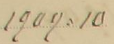, width=2.2cm, height=0.8cm}}            
            \hspace{0.02cm}    
            \subfigure[1907 to 11907] {\epsfig {file=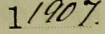, width=2.1cm, height=0.8cm}}
            \hspace{0.02cm}    
            \subfigure[1892 to 92] {\epsfig {file=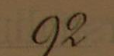, width=1.9cm, height=0.8cm}}            
            \caption{Types of missed labels that were rectified in ARDIS dataset.}
            \label{ardis-labels:fig}

        \end{center}
        \vspace{-2mm}
    \end{figure}

    \textbf{Data annotation.} Data was initially annotated w.r.t. all the locations (bounding boxes) of single digits into the strings of Dataset I since these samples have particularities that are difficult to reproduce synthetically and also there is no information about authors. 
    Furthermore, we have corrected 81 missing labels. 
    A few of them have a different string size (!=4). 
    Figure \ref{ardis-labels:fig} exemplifies samples that had their labels rectified\footnote{The annotated dataset is available upon request for research purposes at https://web.inf.ufpr.br/vri/databases-software/touching-digits}.
    
    \textbf{Experimental Protocol.} We have used the information about the city of origin to split the dataset.
    The samples of a city that were selected for the training set are not present in the test set. 
    The rationale behind this strategy is to evaluate if the model has the ability to generalize the backgrounds since a book and authors will be present in only one of the datasets. Thus, we split the 120 cities into 70\% for training and 30\% for testing, which results in 7,055 samples and 2,945 samples for each set, respectively.

    \textbf{Results.} The performance of end-to-end approach and the segmentation-free approach \cite{Hochuli2018} is presented on Table \ref{ardis-string:tab}.

    We also presented a model named YoLo-TF-Orand, in which we apply the transfer-learning from the model constructed for the Orand dataset (Section \ref{orand-car:sec}). Since both data have a hard background to encode, the rationale here is to evaluate if the acquired learning in one dataset is valuable in another dataset. Thus, we fine-tunned the weights providing historical digit strings from the ARDIS dataset. 
    
\begin{figure}[!h]
        \begin{center}
            \subfigure[] {\epsfig {file=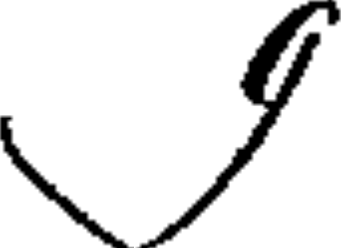, height=0.6cm}}
            \hspace{0.5cm}    
            \subfigure[] {\epsfig {file=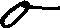, height=0.6cm}}
            \hspace{0.5cm}
            \subfigure[] {\epsfig {file=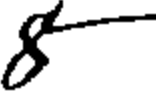, height=0.6cm}}
            \caption{Samples that had their length missed by $\mathcal{L}$ of Hochuli et al.\cite{Hochuli2018}.}
            \label{ardis-missed-length:fig}            
        \end{center}
        \vspace{-5mm}	
    \end{figure}

    As expected, Hochuli et al.\cite{Hochuli2018} suffer from the detection of connected components since it is difficult to define a heuristic that covers all the variety for the background suppression. 
    Moreover, the high variance in handwritten style impact the performance of the length classifier ($\mathcal{L}$) severely, thus misleading the digit classifiers by attributing an isolated digit to a touching digit class (length > 1). 
    Figure \ref{ardis-missed-length:fig} illustrates some of missed length predictions. 
    \vspace{-3.5mm}
    \begin{table}[!h]
        \caption {Comparison of the Accuracy (\%) on ARDIS Numerical Strings (Dataset I)}
        \begin{center}
            \scalebox{0.85}{
                \begin{tabular}{ccc}
                    \hline
                    Method & Accuracy (\%) \\ \hline
                    Hochuli et al.\cite{Hochuli2018} & 25.46 \\             
                    YoLo & 96.80 \\        
                    YoLo-TF-Orand & 95.78 \\ 
                    \hline
                \end{tabular}
            }
            \label{ardis-string:tab}    
        \end{center}    
        \vspace{-3.5mm}
    \end{table}
    
\begin{figure}[!h]
        \begin{center}
            
            \subfigure[] {\epsfig {file=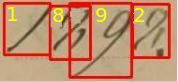,  height=1cm}}
            \hspace{0.1cm}
            \subfigure[] {\epsfig {file=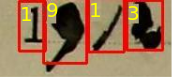, height=1cm}}            
            \hspace{0.1cm}
            \subfigure[] {\epsfig {file=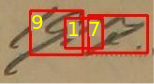, height=1cm}}
            \caption{Missed predictions of YoLo for ORAND dataset: (a) `1898' as `1892', (b) `1912' as `1913',  (c) `1917' as `917'}
            \label{ardis-missed-det:fig}            
        \end{center}
        \vspace{-5mm}
    \end{figure}

    The outstanding performance of YoLo corroborates to the robustness presented in Section \ref{orand-car:sec} in the efficiently encoding of background information. 
    An analysis reveals that most of error sources are in handwriting styles that are poorly represented in dataset (Figure \ref{ardis-missed-det:fig}a), noise samples (Figure \ref{ardis-missed-det:fig}b) and highly slanted numerals (Figure \ref{ardis-missed-det:fig}c).
        
\section{Conclusion}\label{conclusions:sec}

This paper described an end-to-end solution for handwritten numeral string recognition in which the numeral string is assumed to be composed of objects that can be automatically detected and recognized. 
For this purpose, the YoLo architecture \cite{YOLO2016} was used to detect and recognize the digits (objects) inside the string, avoiding heuristic-based methods for string preprocessing and segmentation, or even the need of task-oriented classifiers, and the use of specific constraints related to the string length.

A robust experimental protocol based on three real-world numeral string datasets was used to validate the proposed method: (a) the numeral strings available on the NIST SD19 which are composed of 2- to 6-digits, (b) the courtesy amount of bank check of ORAND dataset and, (c) the recent released ARDIS dataset composed of 4-digit numerals extract from historical documents.
The experimental results have shown that the proposed method is a feasible solution that compares favorably in terms of accuracy to the state-of-the-art in handwritten numeral string recognition. 
Furthermore, the proposed approach performs well on long strings composed of up to 20 digits.

\section*{Acknowledgements}

This research has been supported by CAPES (PhD scholarship - Finance Code 001). In addition, we gratefully acknowledge the support of NVIDIA Corporation with the donation of the TITAN X and V GPUs used for this research.

\balance
\bibliographystyle{IEEEtran}
\bibliography{refer1}
\balance
\end{document}